\def\year{2021}\relax
\documentclass[letterpaper]{article} % DO NOT CHANGE THIS
\usepackage{aaai}  % DO NOT CHANGE THIS
\usepackage{times}  % DO NOT CHANGE THIS
\usepackage{helvet} % DO NOT CHANGE THIS
\usepackage{courier}  % DO NOT CHANGE THIS
\usepackage[hyphens]{url}  % DO NOT CHANGE THIS
\usepackage{graphicx} % DO NOT CHANGE THIS
\usepackage{epsfig}
\usepackage{multirow}
\usepackage{algorithmic}
\usepackage{algorithm}
\usepackage{arydshln}
\usepackage{svg}
\usepackage{amsmath,amssymb}

\usepackage{natbib}  % DO NOT CHANGE THIS AND DO NOT ADD ANY OPTIONS TO IT
\usepackage{caption} % DO NOT CHANGE THIS AND DO NOT ADD ANY OPTIONS TO IT
\usepackage[inline]{enumitem}
\usepackage{comment} 
\usepackage{amsmath}
\newcommand\norm[1]{\left\lVert#1\right\rVert}
\usepackage{todonotes}
\usepackage{mathtools}
\newcommand{\authnoteTM}[2]{{\bf \textcolor{magenta}{#1}: \em \textcolor{magenta}{#2}}}
\newcommand{\tm}[1]{\authnoteTM{TM}{#1}}

\begin{document}
 
\DeclarePairedDelimiterX{\infdivx}[2]{(}{)}{%
  #1\;\delimsize\|\;#2%
}
\newcommand{\infdiv}{\infdivx}
%\DeclarePairedDelimiter{\norm}{\lVert}{\rVert}
\title{Robust and Resource-Efficient Data-Free Knowledge Distillation \\ by Generative Pseudo Replay}
\begin{comment}
\author{
  Paper ID 5745
}
\end{comment}

\author{
  Kuluhan Binici\textsuperscript{\rm 1, 2}, Shivam Aggarwal\textsuperscript{\rm 2}, Nam Trung Pham\textsuperscript{\rm 1}, 
  Karianto Leman\textsuperscript{\rm 1},
  Tulika Mitra\textsuperscript{\rm 2} 
}
\affiliations{
    %Afiliations
    \textsuperscript{\rm 1}Institute for Infocomm Research, A*STAR, Singapore \\
    \textsuperscript{\rm 2}School of Computing, National University of Singapore\\
    %If you have multiple authors and multiple affiliations
    % use superscripts in text and roman font to identify them.
    %For example,

    % Sunil Issar, \textsuperscript{\rm 2}
    % J. Scott Penberthy, \textsuperscript{\rm 3}
    % George Ferguson,\textsuperscript{\rm 4}
    % Hans Guesgen, \textsuperscript{\rm 5}.
    % Note that the comma should be placed BEFORE the superscript for optimum readability
    % email address must be in roman text type, not monospace or sans serif

    % See more examples next
}

\maketitle
\begin{abstract}
\begin{quote}
%\tm{abstract does not use the terms from the title. I have used robust and resource-efficint. But generative pseudo replay is still missing}
 Data-Free Knowledge Distillation (KD) allows knowledge transfer from a trained neural network (teacher) to 
 a more compact 
 %another 
 one (student) in the absence of original training data. Existing works use a validation set to monitor
the accuracy of the student over real data and report
the highest performance throughout the entire process. %a validation set with real data to monitor the widely fluctuating accuracy of the student over time and report the best accuracy during the entire distillation process.
 However, validation data may not be available 
 %in data-free KD using synthetic data, 
 at distillation time either, making it infeasible to record the student snapshot that achieved the peak accuracy.
 %determine the peak accuracy during the distillation process.  
 Therefore, a practical data-free KD method should be robust and ideally provide monotonically increasing student accuracy during distillation. This is challenging because the student experiences knowledge degradation due to the distribution shift of the synthetic data.
 A straightforward approach to overcome this issue is to store and rehearse the generated samples periodically, which increases the memory footprint and creates privacy concerns. We propose to model the distribution of the previously observed synthetic samples with a generative network. In particular, we design a Variational Autoencoder (VAE) with a training objective that is customized to learn the synthetic data representations optimally. 
 The student is rehearsed by the generative pseudo replay technique, with samples produced by the VAE.
 %in conjunction with the newly generated synthetic data.
 Hence knowledge degradation can be prevented without storing any samples. %This technique is called generative pseudo replay. 
 Experiments on image classification benchmarks show that our method optimizes the expected value of the distilled model accuracy while eliminating the large memory overhead incurred by the sample-storing methods. 
\end{quote}
\end{abstract}

\vspace{-10pt}
\section{Introduction}
Recently there is a surging interest to deploy neural networks on the edge devices. Most of these devices have strict resource (e.g., memory or power) constraints that are incompatible with the high computational demand of the neural networks. Knowledge Distillation (KD) \cite{hinton2015distilling} is a well-studied approach to deploy pre-trained neural networks  %(called teacher model) 
on resource-constrained devices by making them more compact. 
%(called student model). \tm{cite KD}  
KD needs access to the original dataset that was used to train the teacher model. This is challenging if the distillation is performed by a party 
%somebody 
other than the %original 
model developers as they may not have access to the original training dataset, either due to privacy issues or the extremely large dataset size making its relocation infeasible.
\par
\begin{figure}[t!]
    \centering
    {\includegraphics[height=.33\textwidth]{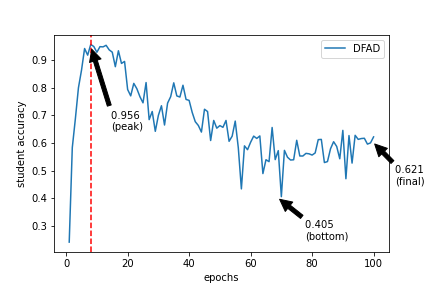}}\hfill

    \caption{Example of student accuracy degradation over distillation steps due to catastrophic forgetting. Red vertical line marks the epoch with peak accuracy on the validation dataset.}
    \label{fig:cat_forget}
    \vspace{-18pt}
\end{figure}

Addressing this issue, some works utilize alternative data that are publicly available ~\cite{addepalli2020degan,nayak2021effectiveness}, while others propose to use synthetic data. ~\cite{zhang2021data, ma2020adversarial}
The second approach is called Data-Free KD and completely eliminates the need for any real data. 
The decoupling of the KD from the original dataset allows distillation to be performed in a much wider spectrum of scenarios in comparison to the data-dependent approaches.
Typically data-free KD approaches contain multiple rounds (epochs)
of synthetic data samples generation and knowledge transfer. In each round, the synthetic samples are generated to close the current information gap between the teacher and the student. As the student is trained and the gap gets smaller, the synthetic sample distribution changes as well.
Therefore, if the previously generated synthetic samples are not periodically rehearsed to the student, the information acquired in earlier distillation epochs might be lost, causing accuracy degradation over time. This is known as catastrophic forgetting \cite{french1999catastrophic} %\tm{cite}. 
Such accuracy degradation is not desirable in practice, especially if the validation dataset is also not available.
%to the user performing the KD, a common scenario. 
The absence of the validation dataset prevents the user from monitoring the student accuracy over time and choosing the distilled model with the peak accuracy. For example, Figure \ref{fig:cat_forget} shows the CIFAR10 validation accuracy of the student model distilled by an existing work, DFAD \cite{fang2019data}, over 100 epochs.  %student accuracy on validation dataset over hundred epochs for \tm{which dataset and which KD method}. 
%in \tm{which epoch}
The peak accuracy is achieved at the $8^{th}$ epoch
and the accuracy degrades subsequently. 
As the users do not have access to the validation dataset, they cannot evaluate the student accuracy and find the peak. Hence, the final accuracy will be a function of the arbitrary termination epoch chosen by the user. This suggests that an ideal data-free KD method should be robust and either sustain high accuracy or monotonically increase it over time so that the distillation can be safely terminated after a pre-determined sufficient number of epochs. 
%\tm{Figure 1 days DFAD but you never explained what it DFAD}
Recent research~\cite{yin2020dreaming, fang2021contrastive} propose to store the generated samples over epochs and continuously expose the student to the entire collection at each distillation step. However, with this approach, the memory overhead and the wall-clock time per distillation step increase substantially. If the dataset is complex in terms of the number of classes and the samples it contains, the required number of distillation steps to achieve an accurate student model can be quite large. For instance, for the Imagenet dataset, the DeepInversion method \cite{yin2020dreaming} stores 140K synthetic samples that occupy around 17 GB and takes 2.8K NVIDIA V100 GPU-hours to complete distillation. Moreover, storing the generated samples may violate the privacy considerations of the original dataset as they can leak information related to the original samples~\cite{li2020membership}.
In summary, the current data-free KD methods are either not robust in terms of student accuracy throughout the distillation process or are resource inefficient by storing the generated samples and can have potential privacy concerns. This trade-off casts them less appealing to the end-users. 
% To democratize data-free KD by allowing any user with a reasonable amount of resources benefit from it, in this work, we propose a novel method that performs rehearsal without storing any samples. 

In this work, we aim to democratize data-free KD by proposing a novel method that preserves student accuracy by performing replay without storing any data samples. We achieve this by designing a  Variational Auto-Encoder (VAE) \cite{kingma2013auto} to model the cumulative distribution of all the generated samples. We establish that the vanilla VAE optimization objective is not suitable to model the synthetic data distribution and propose a synthetic data-aware reconstruction loss. %\tm{add one sentence of how VAE is difficult for synthetic data.}
The modeled distribution is used to infer ``memory samples" that represent those generated in earlier steps and rehearse the student with these samples in conjunction with the newly generated samples to further bridge the information gap between the teacher and the student. Hence, our approach has a constant and very limited memory overhead of a few megabytes to store only the VAE parameters and offers accurate distillation. Experimental results show that our method achieves up to $26.8\%$ increase in average student accuracy compared to methods without replay on image classification benchmarks. Moreover, compared to methods that replay stored samples, ours can reduce the memory footprint by gigabytes. Our concrete contributions can be summarized as follows.  %\tm{offer a highlight of the results.}
%While mitigating the reliability problem in such way, our method only brings constant memory overhead that is limited to the size of VAE parameters.
%This approach mitigates the reliability problem by preventing major decreases in student accuracy throughout distillation, our method introduces constant memory overhead 
%put forward three main contributions:
%To allow any user with reasonable amount of resources benefit from data-free distillation, we tackle the above-mentioned trade-off between resilience against catastrophic forgetting and resource requirement.
%To break the above-mentioned trade-off between resilience against catastrophic forgetting and distillation speed, in this work, we put forward three main contributions:
\begin{itemize}
    \item A novel data-free KD framework containing two specialized generators to allow the student network acquire new knowledge while retaining the prior information.
    \item Enabling VAEs to operate with synthetic data, produced for knowledge distillation, by modifying the reconstruction loss.
    \item Extensive experimental evaluation of our approach in comparison with the state-of-the-art.
\end{itemize}

\begin{figure*}[h!]
        %{\includegraphics[height=.28\textwidth]{figures/KD_stage.png}}\hfill
        %{\includegraphics[height=.28\textwidth]{figures/gen_stage.png}}\hfill
        \centering
        {\includegraphics[height=.55\textwidth]{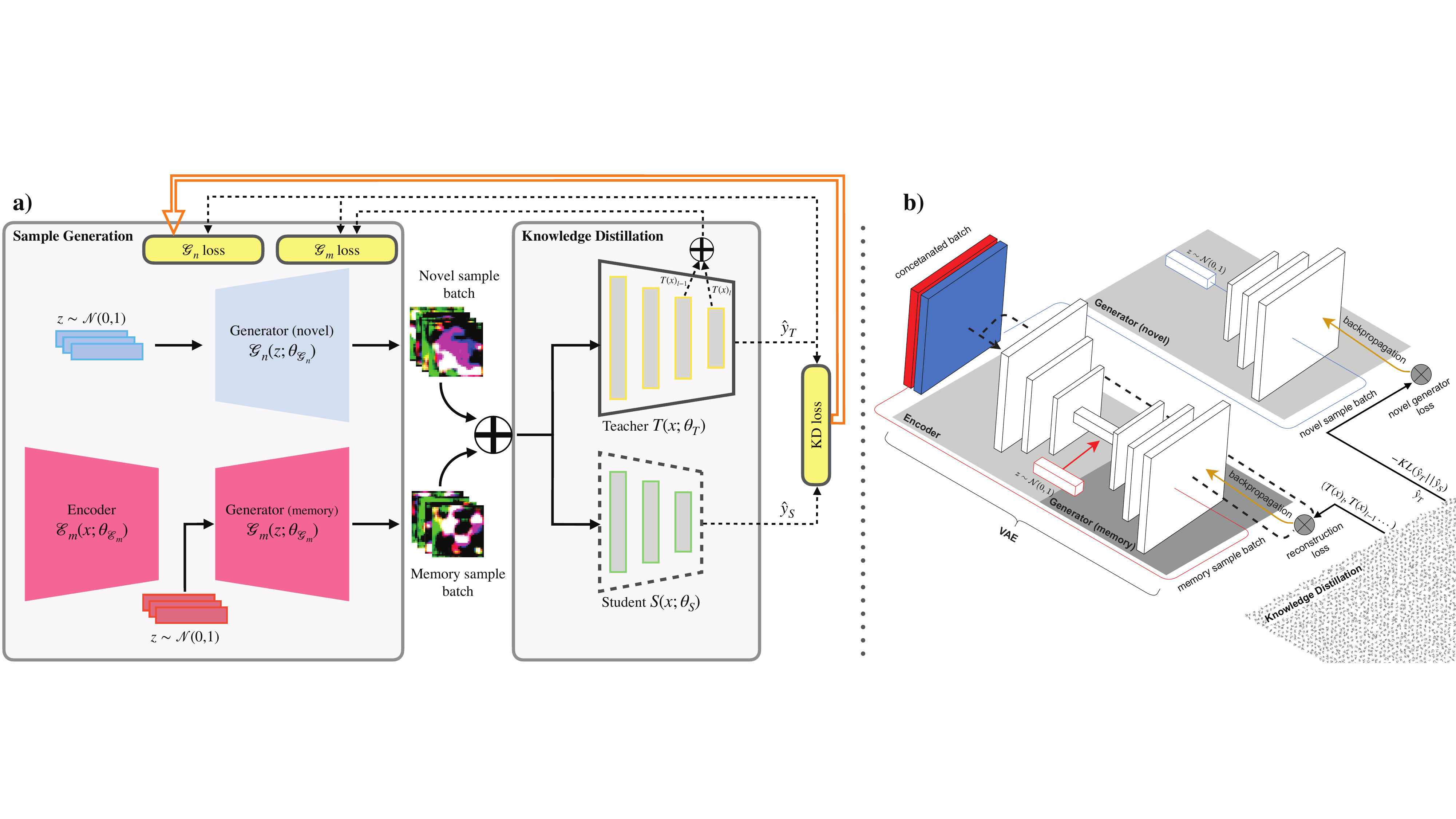}}
        \vspace{-50pt}
        \caption{Method overview. (a) Proposed PRE-DFKD framework. The student and the generators are trained alternately. First the generators are fixed and the student is trained by a combination of novel and memory samples. In the next stage, student is fixed and the generators are trained by the learning signals received from the KD. (b) Training process of VAE. First, the memory generator is frozen and a batch of memory samples are inferred. Later the memory batch is combined with a novel sample batch to train the encoder-decoder pair. }
        \label{fig:method}
        \vspace{-10pt}
    \end{figure*}

\vspace{-10pt}
\section{Related Work}
\subsection{Knowledge Distillation (KD)}
\label{sec:rel-KD}
KD \cite{hinton2015distilling} can be defined as transferring the learnt prediction behavior of a complex neural network model to a relatively smaller one. In literature, the
complex network is often referred as the “teacher” and the compact network is referred as the
“student”. Throughout the distillation process, the student is trained with the guidance of the ground truth
labels as well as the teacher’s responses to the training data. The inclusion of teacher’s responses in the training objective consolidates the information provided to the student about the data-label relationships. Thus the student network can achieve much higher accuracy than when it is trained with the supervision of only the ground truth labels. This is viewed as a form of compression, as the compact student model approximates the teacher by carrying more information than it could have learnt from data on its own. 
% KD has been proven effective in maintaining model performance after compression ~\cite{}. %~\cite{asami2017domain,shi2019knowledge,sun2019patient}. 
%The process can be defined as extracting knowledge from a complex pre-trained network, called the teacher, to train a relatively more compact student.
%This extraction is generally done by forcing the student to imitate the responses of the teacher against the training data. 
\begin{comment}

In vanilla KD introduced by \cite{hinton2015distilling}, the student is trained according to the objective in Eq. \ref{eq8}.
\begin{equation}
\mathcal{L}_{S} = \mathcal{L}_{CE} + \mathcal{L}_{KD}
\label{eq8}
\end{equation}
\begin{equation}
\mathcal{L}_{KD} = \frac{1}{N}\sum_{i=1}^{N}KL(\infdiv{\sigma(\mathcal{T}(x^i), T)}{\sigma(\mathcal{S}(x^i), T)})
\label{eq9}
\end{equation}
\begin{equation}
\sigma(x, T)_i = \frac{e^{x_i/T}}{\sum_{j}e^{x_j/T}}
\label{eq10}
\end{equation}
\begin{equation}
\mathcal{L}_{CE} = \frac{1}{N}\sum_{i=1}^{N}\mathcal{H}(y^i, \mathcal{S}(x^i))
\label{eq11}
\end{equation}
The $\mathcal{L}_{KD}$ in Eq. \ref{eq9} is the distillation loss measured by the KL-divergence between teacher and student predictions. The $\mathcal{L}_{S}$ in Eq. \ref{eq11} is the cross-entropy loss between ground truth labels and student predictions. 
\end{comment}
%Therefore the distillation objective becomes equal to eq. \ref{eq9}.
\vspace{-5pt}
\subsection{Data-Free KD}
\label{sec:rel-DFKD}
The concept of performing knowledge distillation using synthetic data samples in place of a real dataset is called data-free KD. Unavailability of real data with ground truth labels limits the student training to be guided only by the teacher outputs (softmax logits, activation maps, etc.). One quality that the synthetic samples are expected to have is that their distribution should match that of the original data ~\cite{yoo2019knowledge, nayak2019zero, chen2019data, haroush2020knowledge}. Another one is that they should be optimal to close the information gap between the teacher and the student ~\cite{micaelli2019zero, fang2019data}. Producing samples that satisfy both, generally yields the best distillation results ~\cite{yin2020dreaming, fang2021contrastive, binici2021preventing}. During distillation, the synthetic data distribution is updated at each epoch to achieve the above-mentioned qualities. Such distribution shift could cause the student model to lose the information over epochs and experience accuracy degradation \cite{binici2021preventing}. This could be avoided by periodically rehearsing the student with samples from all the distributions it has previously observed. We can group the existing data-free KD methods based on whether they employ such practice or not as {\em replay-based methods} and {\em replay-free methods}.
\par
The replay-based methods typically store a collection of generated synthetic samples over epochs. CMI \cite{fang2021contrastive} is a recent work that utilizes the model inversion technique to generate the synthetic samples and store them all in memory.  Each time a newly generated batch is added, the student is trained with the entire collection of samples. This approach suffers from high memory utilization and requires a significant amount of time to achieve high-quality distillation.  \cite{binici2021preventing} proposed to use a fixed-sized memory buffer instead of storing all synthetic samples. Although the memory overhead is constant, this approach does not completely eliminate the possibility of catastrophic forgetting as the student is rehearsed only with a subset of previously observed samples.
\par
The replay-free approaches train the student only with the newly generated samples at any distillation epoch. DAFL \cite{chen2019data} introduced three novel loss terms that motivate the generated samples to be categorically diverse and be classified with high confidence by the teacher. DFAD \cite{fang2019data} targeted generating samples that would cause maximum information gain to the student when learned. 
%The and DAFD \cite{fang2019data} are popular works from this category and use generative neural networks to produce synthetic samples. %
Since these works do not contain any replay mechanism, the student accuracy often fluctuates and degrades over epochs.

\subsection{Replay in Continual Learning}
Continual Learning (CL) methods focus on incrementally learning from a non-stationary dataset, with the goal of mitigating catastrophic forgetting \cite{goodfellow2015empirical}, i.e., the inability of the model to preserve the past knowledge while learning from the recent data. The CL methods \cite{mai2021online} can be broadly classified into three major approaches: regularization-based, architecture-based, and replay-based (either generative or from a stored buffer), with the latter being the most effective and the focus of this paper.
% \tm{is the last part of the sentence applies only to replay-based? Then it should be in brackets}. \tm{add one sentence on why you focus on replay-based methods.} 
Vanilla replay-based techniques~\cite{Rebuffi2017iCaRLIC} maintain a buffer of raw samples from the past tasks and replay them periodically with the new data in the learning process. However, these methods become infeasible when raw samples cannot be stored due to privacy or memory constraints. Moreover, they cannot perform lifelong learning due to the inability to scale well with the growing data stream. The generative replay strategies~\cite{shin2017continual, DBLP:journals/corr/abs-1802-00853} mitigate these shortcomings by training a deep generative model to create pseudo-samples that mimic the data distribution from past experiences. Our work leverages generative pseudo replay. In particular, we adopt a VAE-based strategy \cite{shin2017continual} to preserve the knowledge of the previously generated synthetic samples. Unlike existing continual learning works, our method focuses on the generation of synthetic samples in a data-free environment. 
% This motivates the need to tackle the incompatibility of the with the synthetic data \tm{unclear}. 
We propose a modified optimization objective to preserve the categorical information in reconstructed samples and mitigate catastrophic forgetting in a data-free knowledge distillation scenario.

\label{sec:rel-pseudo}
\section{Proposed PRE-DFKD Approach}
\label{sec:method}
Our data-free KD approach is called \textit{Pseudo Replay Enhanced Data-Free Knowledge Distillation} (PRE-DFKD). Figure \ref{fig:method}(a) provides an overview of PRE-DFKD. It consists of {\em data generation} and {\em knowledge distillation}. In data generation, randomly sampled latent variables are transformed by generative models to produce synthetic samples. Later during knowledge distillation, the student model is trained to categorize these samples similar to the teacher model. These two stages repeat alternately until the target number of steps is reached.  We use two generative models in our framework. The first one ({\em novel sample generator)} produces samples that bring novel information to the student. The second generator ({\em memory sample generator}) is responsible for re-exposing the student to the information acquired earlier.
\subsection{Novel Sample Generation}
We define novel samples as those that the student classifies differently from the teacher. 
%The samples that cause the largest disagreement between these two models are considered to be most novel. 
When the student is trained with these novel samples, it better approximates the teacher. We condition our novel sample generator by including the %KLD 
 distance between student and teacher predictions in the optimization objective. To quantify such distance, we used Jensen-Shannon (JS) divergence (see Eq. \ref{eq3})~\cite{yin2020dreaming} as it is shown to provide better performing student models compared to other alternatives (e.g. Kullback–Leibler, L1 norm)  \cite{binici2021preventing}.
Moreover, to ensure that the novel sample distribution is similar to the original one, we also include the loss terms introduced by \cite{chen2019data} (see Eq. \ref{eq2}). The first two of these are {\em predictive entropy} and {\em activation} loss terms, which are minimized when the synthetic data induce the teacher to output high valued activation maps and low entropy prediction vectors. The third term, {\em categorical entropy} loss, sustains class balance in the generated batches by maximizing the categorical entropy of synthetic sample distribution.
%These terms are minimized when the synthetic data induce the teacher to output high valued activation maps and low entropy prediction vectors. 
%Lastly, we consider the categorical distribution of the synthetic samples and maximize its entropy to sustain class balance in the generated batches. 
Eq. \ref{eq1} shows the complete optimization objective of the novel sample generator.
\begin{equation}
\theta_{n}^{*} := \arg\min\limits_{\theta}\left(\mathcal{L}^{(T)}_{\phi} + \alpha \mathcal{L}_{\delta}\right)
\label{eq1}
\end{equation}
\begin{equation}
\mathcal{L}^{(T)}_{\phi} = \frac{1}{n} \sum_i \left( \lambda_1t_{T}^ilog(\hat y_{T}^i) - \lambda_2\norm{f_{T}^i}_1 \right)
-\lambda_3\mathcal{H}(p(\hat y_{T}))
\label{eq2}
\end{equation}
\begin{equation}
%\mathcal{L}_{\delta} = -KL\infdiv{\hat y_{T}}{\hat y_{S}}; \quad z \sim \mathcal{N}(0,1)
\mathcal{L}_{\delta} = 1-JS\infdiv{\hat y_{T}}{\hat y_{S}}; \quad z \sim \mathcal{N}(0,1)
\label{eq3}
\end{equation}
where $\theta_{n}^{*}$, represents the optimal values of the generator parameters denoted by $\theta_{n}$. Moreover, $\hat y_{T}=\mathcal{T}(\theta(z))$ and $f_{T}$ are the softmax outputs and the activation maps at the last FC layer of the teacher $\mathcal{T}$ for the generated batch $\theta(z)$ respectively, while $t_{T}^i = argmax(\hat y_{T}^i)$. $\mathcal{H}(p(\hat y_{T}))$ denotes the entropy of class label distribution in generated batch. $\lambda_i (i=1,2,3)$ and $\alpha$ coefficients adjust the weighted contribution of each loss term.
%\tm{you did not talk about generated batches before}
\vspace{-2pt}
\subsection{Memory Sample Generation}
\label{par:memory_gen}
\begin{figure}[t!]
    \centering
    \includegraphics[height=.170\textwidth]{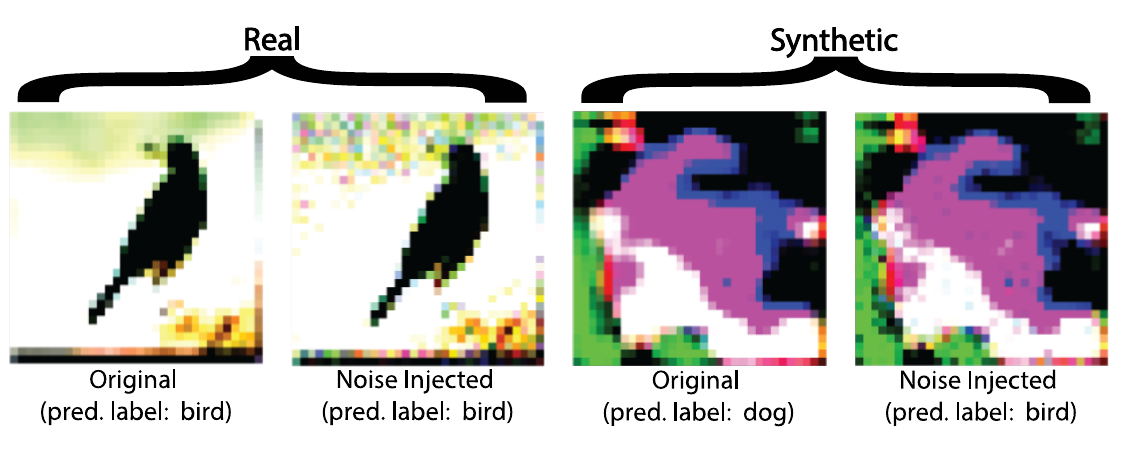}
    \includegraphics[height=.205\textwidth]{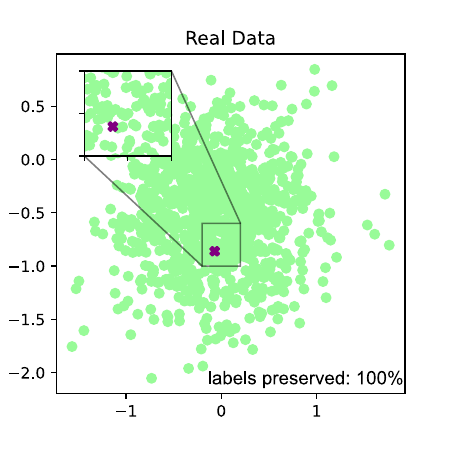}
    \includegraphics[height=.205\textwidth]{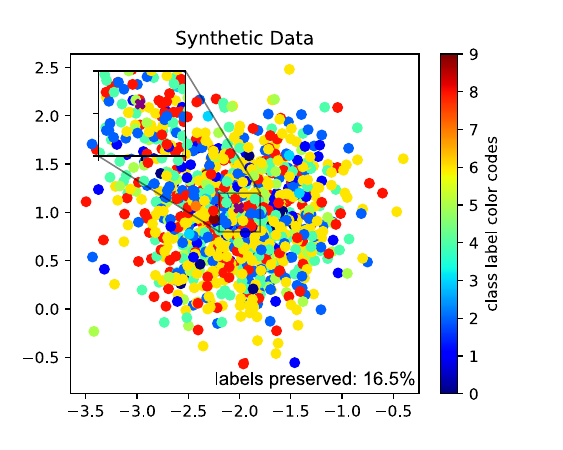}
    %\vspace{-5pt}
    \caption{Example real and synthetic CIFAR10 samples are given before and after noise injection in the first row.  In the second row, the projection of real and synthetic samples injected with noise on 2-dimensional planes are shown. Colors represent the class labels assigned to these samples by the teacher model. The original samples are marked with purple crosses.}
    \label{fig:noise_experiment}
\end{figure}

 As mentioned earlier, the distribution of novel samples shifts over distillation steps. To prevent such shift from causing the student to lose earlier learned information, we replay the samples from earlier distributions. These memory samples are inferred from a VAE that models the past synthetic data distributions. Our memory sample generator is the decoder part of the VAE and is trained jointly with the encoder. To train the pair, we use a combination of novel samples and memory samples. The inclusion of memory samples in the training process is to prevent the generator itself from experiencing catastrophic forgetting. The process is described in Algorithm \ref{alg:mem_gen_train} and visualized in Figure \ref{fig:method}(b). %\tm{for readers who don't know, you need to explain why you need a pair. Also add a paragraph on Challenges in Memory sample generations before talking about the solutions}
 \setlength{\textfloatsep}{10pt}
 \begin{algorithm}[!h]
\caption{Memory Sample Generator Training}
\label{alg:mem_gen_train}
\begin{algorithmic}
\STATE \text{\textbf{INPUT:}} Novel sample generator $\mathcal{G}_n(z; \theta_{\mathcal{G}_n})$, memory sample generator $\mathcal{G}_m(z; \theta_{\mathcal{G}_m})$, encoder $\mathcal{E}_m(x; \theta_{\mathcal{E}_m})$, batch size b, latent vector dimension n.
\\[5pt]
%\newline
\STATE \# Train with novel samples \
\STATE sample B vectors ($z \sim \mathcal{N}(0,1)$)\
\STATE $x_{n} \gets \mathcal{G}_n(z)$\
\STATE $\hat{z}_{\mu}, \hat{z}_{\sigma} \gets \mathcal{E}_m(x_{n})$\
\STATE $z_{n} \sim \mathcal{N}(\hat{z}_{\mu},\hat{z}_{\sigma})$\
\STATE $\hat{x}_{n} \gets \mathcal{G}_m(z_{n})$\
%\newline
\\[5pt]
\STATE \# Rehearse by reconstructing old samples \
\STATE sample B vectors ($z_m \sim \mathcal{N}(0,1)$)\
\STATE $x_{m} \gets \mathcal{G}_m(z_m)$\
\STATE $\bar{z}_{\mu}, \bar{z}_{\sigma} \gets \mathcal{E}_m(x_{m})$\
\STATE $\bar{z}_{m} \sim \mathcal{N}(\bar{z}_{\mu},\bar{z}_{\sigma})$\
\STATE $\hat{x}_{m} \gets \mathcal{G}_m(\bar{z}_{m})$\
\\[5pt]
%\newline
\STATE \# Calculate \& backpropagate loss \
\STATE $\mathcal{L}_{VAE} \gets \mathcal{L}_{rec}(x_{m}^{(i)}, \hat{x}_{m}^{(i)}, x_{n}^{(i)}, \hat{x}_{n}^{(i)}) + \gamma \mathcal{L}_{KLD}(\bar{z}_{\mu},\bar{z}_{\sigma})$
\STATE $\theta_{\mathcal{G}_m}, \theta_{\mathcal{E}_m} \gets optimizer.step(backward(\mathcal{L}_{VAE}),\theta_{\mathcal{G}_m}, \theta_{\mathcal{E}_m})$
\end{algorithmic}
\end{algorithm}
%\vspace{-10pt}
%\paragraph{VAE Reconstruction Loss for Synthetic Training Data}
\vspace{-10pt}
\paragraph{Incompatibility of Vanilla VAE loss for Synthetic Data Representation Learning}
After conducting experiments with vanilla VAE for pseudo replay in data-free KD, we observed that the teacher's predictions of the memory samples were of low confidence (14\%). However, the average prediction confidence was significantly higher for novel samples (72\%). This indicated that the modelled distribution was not accurate. We posited that the training loss of the vanilla VAE is incompatible with the properties of synthetic samples. The reconstruction term in the loss function ($\mathcal{L}_{rec} = \norm{x - D(E(x))}_2^2$) is optimized when the decoded samples $D(E(x))$ match the inputs $x$ in pixel space. If the L2 distance between the input and target image pixels is low, the loss term yields a small value. This assumes that small differences in pixel values do not impact the information content significantly. While this assumption holds for real images, it might not be accurate for synthetic ones. 
%\tm{text fonts in your figure are somewhat blurred. I think you are not using embedded fonts. Also put real images on the left to match top and bottom figures}
The synthetic samples are generated for the sole purpose of knowledge transfer. They are not visually realistic (see Figure \ref{fig:noise_experiment}) yet still achieve their goal. Therefore, the assumption that two images that are visually similar represent the same content may not hold for synthetic samples. %\tm{explain bottom figure. In fact, the bottom figure should be on top.}
To test this hypothesis, we inject random Gaussian noise to both real and synthetic samples and record the percentage of samples that preserved their class labels. The experimental details are given in the appendix. 
The effect of pixel-wise perturbations on class labels is visualized in Figure \ref{fig:noise_experiment}. It can be seen that the class information of synthetic samples got affected more significantly than real samples. This suggests that even if a reconstructed sample yields a small reconstruction loss with very similar pixel values to those of the synthetic input, its categorical information could be completely different. 
\vspace{-23pt}
\paragraph{Synthetic Data-Aware VAE Reconstruction Loss}
To ensure that the categorical information is preserved, we modify the VAE reconstruction loss. Our synthetic data-aware loss constrains the reconstructed samples to cause similar feature responses in the last layers of the teacher. Typically, these layers are considered to represent content-related information in Convolutional Neural Networks (CNNs) \cite{jing2019neural}. To impose the mentioned constraint, we add the term in Eq. \ref{eq4} to the VAE optimization objective.
%to account for the categorical information of the target samples interpreted by the teacher. \todo{rephrase this sentence}. 
\begin{equation}
\sum_{l\in L}|T(x)_l - T(D(E(x)))_l|_1
\label{eq4}
\end{equation}
The encoder and decoder networks are represented by $D$ and $E$ respectively. %\tm{overloading of variable E as you use it for expectation as well}. 
Moreover, the set $L$ denotes the selected set of teacher network ($T$) layers. The total training loss we use can be denoted as, 
\begin{equation}
\mathcal{L}_{VAE} = \mathcal{L}_{rec} + \mathcal{L}_{KLD}
\label{eq5}
\end{equation}
\begin{equation}
\mathcal{L}_{rec} = |x - D(E(x))|_1 + \sum_{l\in L}|T(x)_l - T(D(E(x)))_l|_1
\label{eq6}
\end{equation}
\begin{equation}
\mathcal{L}_{KLD} = \infdiv{\mathcal{N}(\mu_z,\sigma_z)}{\mathcal{N}(0,1)}
\label{eq7}
\end{equation}
where $x$ is the input sample batch; $\mu_z$ and $\sigma_z$ are the mean and variance of the latent vector ($z=E(x)$) distribution. 

\paragraph{Sustaining Class Balance in Inferred Memory Batches} Once we accomplish training a VAE with synthetic data, another remaining issue is to ensure that the inferred memory samples are evenly distributed into classes. %\tm{why do you need to do this?}. 
%This is to ensure that the student retains knowledge related to every available class. %memory batch used to rehearse the student is not biased  
Typically, latent vectors that are sampled from standard Gaussian distribution are fed to the generator to produce new images. However, our task requires the generated batch to be diverse in terms of the categorical contents of the samples. This is to guarantee that the student learns to approximate the teacher for samples from any category. Therefore, to provide such class balance, we freeze the memory sample generator and tune the input latent vectors. 
\begin{equation}
z_m^* := \arg\max\limits_{z~\sim \mathcal{N}(0,1)}\left(p(\hat y_{N})log(p(\hat y_{N}))\right)
\label{eq8}
\end{equation}
The procedure we used to solve the optimization objective described in Eq. \ref{eq8} is given in algorithm \ref{alg:mem_gen_infer}.
\begin{algorithm}[!h]
\caption{Memory Sample Generator Inference}
\label{alg:mem_gen_infer}
\begin{algorithmic}
\STATE \text{\textbf{INPUT:}} memory sample generator $\mathcal{G}_m(z; \theta_{\mathcal{G}_m})$, teacher $T(x;\theta_T)$.
\STATE \text{\textbf{OUTPUT:}} memory batch $x_{m}$.
\newline
\STATE \# Tune latent variables \
\STATE sample B vectors ($z_m \sim \mathcal{N}(0,1)$)\
\STATE $x_{m} \gets \mathcal{G}_m(z_m)$\
\STATE $\hat{y}_{\mathcal{T}} \gets \mathcal{T}(x_{m})$\
\STATE $\mathcal{R} \gets \frac{1}{n} \sum_i \left(t_{T}^ilog(\hat y_{T}^i)\right) +KL\infdiv{\mathcal{N}(\mu_z,\sigma_z)}{\mathcal{N}(0,1)}$\
\STATE $\mathcal{L}_z \gets - \mathcal{H}(p(\hat y_{\mathcal{T}})) + \mathcal{R}$
\STATE $z_m \gets optimizer.step(backward(\mathcal{L}_{z}),z)$
\newline
\STATE \# Infer memory samples \
\STATE $x_{m} \gets \mathcal{G}_m(z_m)$\
\end{algorithmic}
\end{algorithm}
Tuning the input latent variables can deviate their distribution from standard normal, which would cause the generator to produce noise. To prevent such deviation, we use the Kullback–Leibler (KL) divergence as a regularization term. Moreover, to sample the most representative images (within each class) from the modeled distribution, we also include the predictive entropy loss in the regularization term ($\mathcal{R}$). %\tm{never mentioned KL divergence before}
%\todo{KLD or KL divergence?} 
\subsection{Knowledge Distillation}
Knowledge transfer happens by minimizing the distance between teacher and student predictions against combined batches of novel and memory samples. The minimization problem that results in optimal student model parameters is
\begin{equation}
%\theta_S^* := \arg\min\limits_{\theta_S}KL\infdiv{S((x_n,x_m);\theta_S)}{T((x_n,x_m);\theta_T)}
\theta_S^* := \arg\min\limits_{\theta_S}\norm{S((x_n,x_m);\theta_S)- T((x_n,x_m);\theta_T)}_1
\label{eq9}
\end{equation}
$\theta_S^*$ in Eq. \ref{eq9} denotes the optimal student model parameters; $x_n$ and $x_m$ are novel and memory samples, respectively.
\begin{table*}[!ht]
\centering
\begin{tabular}{l|l|l|l|l|l|l|l|l|l|l|ll}
\hline
                & \multicolumn{3}{c|}{MNIST}         & \multicolumn{3}{c|}{CIFAR10}          & \multicolumn{3}{c|}{CIFAR100}       & \multicolumn{3}{c}{Tiny ImageNet}                       \\ 
                 & \multicolumn{3}{c|}{$\mathcal{T}:$LeNet5 (98.9\%)}        & \multicolumn{3}{c|}{$\mathcal{T}:$ResNet34 (95.4\%)}        &
                 \multicolumn{3}{c|}{$\mathcal{T}:$ResNet34 (77.9\%)}         &
                 \multicolumn{3}{c}{$\mathcal{T}:$ResNet34 (71.2\%)}  \\
                & \multicolumn{3}{c|}{$\mathcal{S}:$LeNet-half} & \multicolumn{3}{c|}{$\mathcal{S}:$ResNet18} &
                \multicolumn{3}{c|}{$\mathcal{S}:$ResNet18}&
                \multicolumn{3}{c}{$\mathcal{S}:$ResNet18} \\ \hline
\textbf{Method} & \textbf{$\mu$} & \textbf{$\sigma^2$} & \textbf{$acc_{max}$}& \textbf{$\mu$} & \textbf{$\sigma^2$} & \textbf{$acc_{max}$}& \textbf{$\mu$} & \textbf{$\sigma^2$} & \textbf{$acc_{max}$} & \multicolumn{1}{l|}{\textbf{$\mu$}} & \multicolumn{1}{l|}{\textbf{$\sigma^2$}} & \textbf{$acc_{max}$} \\ \hline
Train with data         &   98.7        &  0.5   &    98.9  &    89.0      &      8.1  &  95.2    &   71.3       &  8.1  &    77.1   &   60.2      & \multicolumn{1}{l|}{8.8}    & 64.9        \\ \hline
DAFL            & 87.3          & 6.6     &  98.2     & 62.6          & 17.1      &   92.0     & 52.5          & 12.8      &  74.5     &     39.5     & \multicolumn{1}{l|}{\textbf{10.3}} & 52.2           \\
DFAD            & 63.5          & 6.8     &   98.3    & 86.1          & 12.3      &   93.3     & 54.9          & 12.9      &   67.7    & -         & \multicolumn{1}{l|}{-}   & -         \\ \hline
\multirow{2}{*}{CMI}  &  \multicolumn{3}{c|}{memory: N.A. } &  \multicolumn{3}{c|}{memory: 250 MB }           &  \multicolumn{3}{c|}{memory: 500MB }                 &  \multicolumn{3}{c}{memory: 2.7 GB }                  \\          & -         & -     &    -   &   82.4    &   16.6   &    \textbf{94.8}    &  55.2           & 24.1     &  77.0     & -         & \multicolumn{1}{l|}{-}  & -          \\ \hdashline
\multirow{2}{*}{MB-DFKD}   &  \multicolumn{3}{c|}{memory: 6.7 MB}           &  \multicolumn{3}{c|}{memory: 20 MB}                 &  \multicolumn{3}{c|}{memory: 20 MB}     &  \multicolumn{3}{c}{memory: 20MB}              \\        & 88.6          & 3.2     &   98.3    & 83.3          & 16.4      &    92.4    & 64.4          & 18.3      & 75.4      & 45.7         & \multicolumn{1}{l|}{11.5}  & 53.5          \\ \hdashline
\multirow{2}{*}{PRE-DFKD (ours)}  &  \multicolumn{3}{c|}{memory: 2.1 MB}           &  \multicolumn{3}{c|}{memory: 2.1 MB}                 &  \multicolumn{3}{c|}{memory: 2.1 MB}                &  \multicolumn{3}{c}{memory: 2.1 MB}   \\          & \textbf{90.3}          & \textbf{1.9}     &    \textbf{98.3}   & \textbf{87.4}          & \textbf{10.3}      &   94.1     & \textbf{70.2}          & \textbf{11.1}      &    \textbf{77.1}   & \textbf{46.3}         & \multicolumn{1}{l|}{11.0} & \textbf{54.2}            \\ \hline
\end{tabular}
\caption{Student accuracy results for Data-Free KD on four image classification benchmarks. The $T$ and $S$ values denote the teacher-student pairs. The values for $\mu$ and $\sigma^2$ represent mean and variance $(\%)$ of the averaged student validation accuracy (over 4 runs), over epochs. $acc_{max}$ is the maximum recorded accuracy at any epoch of any distillation run. %\tm{explain up and down arrow}
} 
\label{table:main}
\end{table*}
\section{Experimental Evaluation}
%\tm{you are using replay and rehearsal interchangeably} 
We demonstrate the effectiveness of PRE-DFKD in improving the expected student accuracy and reducing resource utilization on several image classification benchmarks. We provide a comparison with methods that store samples for replay and those that do not employ replay. Our baselines from the first category are CMI~\cite{fang2021contrastive}, and the memory bank approach~\cite{binici2021preventing} that we refer to as MB-DFKD. For the baselines that do not use replay, we select DAFL~\cite{chen2019data} and DFAD~\cite{fang2019data} methods. For a fair comparison, we used the implementations of CMI, DAFL, and DFAD available from the authors' GitHub pages. 
%As the code for MB-DFKD was not available, we implemented the approach based on the paper. 

%Moreover, we conduct an ablation study to verify that each utilized technique regarding pseudo rehearsal has a positive contribution on the results. 
\paragraph{Datasets:}
We use four datasets with different complexities and sizes. The simplest is MNIST \cite{lecun1998gradient} that contains $32\times 32$ grayscale images from ten classes. CIFAR10 \cite{krizhevsky2009learning} contains RGB images from ten classes ($3\times 32\times 32$). CIFAR100 \cite{krizhevsky2009learning} contains hundred different categories while the samples have the same dimensions as those in CIFAR10. Lastly, the most complex dataset we use is Tiny ImageNet \cite{deng2009imagenet} that contains $64\times 64$ RGB samples from 200 classes.
\paragraph{Implementation Details}
We run each method for 200, 200, 400, and 500 epochs for MNIST, CIFAR10, CIFAR100, and Tiny ImageNet, respectively.  To evaluate each dataset and method pair, we conduct four runs. For MNIST, we select LeNet5~\cite{lecun1998gradient} and LeNet5-half as the teacher-student pair. For the remaining datasets, we use ResNet34 \cite{he2016deep} as the teacher and ResNet18 as the student. While benchmarking the baseline methods, we use the hyper-parameter configurations from the papers or GitHub pages where available. Our code and additional experimental details are available at \url{https://github.com/kuluhan/PRE-DFKD}. We note that CMI failed for MNIST and Tiny ImageNet datasets. The original implementation required the teacher models to contain Batch Normalization layers and as LeNet does not contain any, we could not test it on MNIST. For Tiny ImageNet, although we searched extensively for proper hyper-parameter configuration, none of our trials yielded comparable results. Similarly, all our experiments with DFAD on Tiny ImageNet were unsuccessful. 
%\paragraph{Student-Teacher Pair:}
%\tm{choose either past or present tense and stick with it. Don't mix up. I have fixed it here.}
%For MNIST, we select LeNet5~\cite{lecun1998gradient} and LeNet5-half as the teacher-student pair. We use a batch size of 512 for novel samples. To train the student and the generators, we use ADAM optimizer. The learning rates  used in training are given in the appendix \ref{}. For the remaining datasets, we use ResNet34 \cite{he2016deep} as the teacher and ResNet18 as the student. We train the students with SGD optimizer. For CIFAR10/100 datasets we set the batch size for novel samples as 1,024, while for Tiny ImageNet we set it as 1,500. For the baseline methods, we use the hyper-parameter configurations from the papers or GitHub pages where available. For others, we report the best results after trials with several different configurations. 
\vspace{-10pt}
\paragraph{Evaluation Metrics:}
If we consider the termination step of the distillation as a random variable $ts \in \mathbb{R}$, then our goal is
\begin{equation}
\max_{ts} \mathbb{E}[acc_{ts}] - \sigma^2[acc_{ts}] 
\end{equation}
where $acc_{ts}$ stands for student accuracy at epoch $ts$. To report results, first, we average the $acc_{ts}$ across all runs at the corresponding epoch ($ts$). These averaged accuracy series can be denoted as $\mathbb{E}_{i}[acc_{ts}^{(i)}]$ where $acc_{ts}^{(i)}$ is the student accuracy observed in the $i^{th}$ run at epoch $ts$. We report the mean and variance values of $\mathbb{E}_{i}[acc_{ts}^{(i)}]$. The mean ($\mu$) corresponds to $\mathbb{E}_{ts}[\mathbb{E}_{i}[acc_{ts}^{(i)}]]$ and the variance ($\sigma^2$) corresponds to $\sigma^2[\mathbb{E}_{i}[acc_{ts}^{(i)}]]$.
%We report the mean ($\mu$) and variance ($\sigma^2$) to visualize the 
%Therefore, we report the average mean ($\mathbb{E}[.]$ or $\mu$) and variances ($\sigma^2[.]$) of the student accuracy recorded after multiple runs. 
%the $\sigma^2[.]$ considers the deviation of both values that are higher than the mean and those 
%variance quantifies the total deviation from mean regardless
Additionally, we include the peak accuracy recorded across all runs ($acc_{max}$) to demonstrate that our method achieves state-of-the-art performance when the validation set is available. 
%number of epochs at which the student accuracy was lower than the average ($n^{-}_{\mu}$). 

%We For MNIST, we set the number of distillation epochs $n_{epochs}$ as 200 for all methods. The learning rates we used to run PRE-DFKD were 0.002 and 0.2 for the student $\lambda_S$ and novel sample generator $\lambda_{Gn}$ respectively. The novel sample batch size $B_n$ was 512. The learning rate for memory sample generator $\lambda_{Gm}$ was $0.1\lambda_{Gn}$, memory sample batch size $B_m$ was $B_n / 8$ in all benchmarks.

\begin{figure*}[!ht]
\centering
%\centerline{\includesvg[height=.20\textwidth]{figures/MNIST_comparsion.svg}}
\centerline{\includegraphics[height=.20\textwidth]{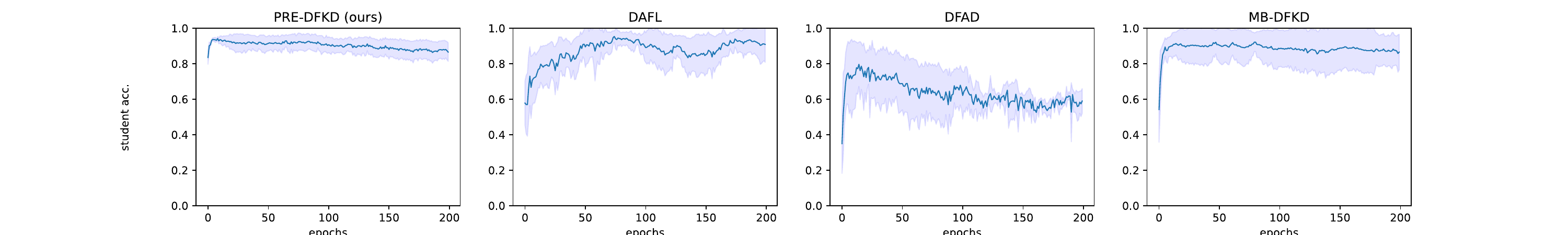}}
%\centerline{(a) Benchmark on MNIST}\smallskip
\centerline{\includegraphics[height=.20\textwidth]{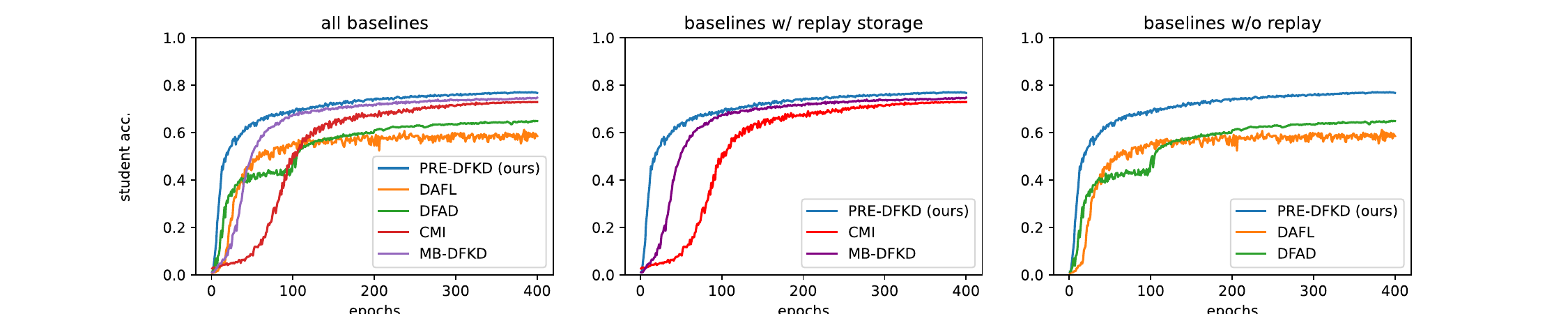}}
%\centerline{(b) Benchmark on CIFAR100}\smallskip
\caption{Visualized student accuracy curves. The first row contains MNIST results with mean and variance values over runs. The second row contains student accuracies averaged over runs for CIFAR100.}
\label{fig:acc_plots}
\end{figure*}

\paragraph{Improved Distillation Quality:} Table \ref{table:main} provides
a summary of the results across four runs. The accuracies of the pre-trained teachers are given next to $\mathcal{T}$ in Table \ref{table:main} followed by the student accuracy with different methods. The first row {\em Train with data} shows the student accuracy when trained with the original dataset. %\tm{why not available for ImageNet?}.
The remaining rows report the accuracy for different baseline data-free KD methods and our approach (PRE-DFKD). As expected, the {\em Train with data} student accuracy is lower than the teacher accuracy and the data-free methods have lower student accuracy than {\em Train with data} student accuracy. PRE-DFKD achieves student accuracy that is better than almost all the baseline data-free KD approaches and approximates %approaches closer to 
the student accuracy trained with the original dataset closer for all datasets. %\tm{you don't want to report peak accuracy anywhere? Might be useful if validation dataset is available.}
When compared with baselines that do not contain replay (DAFL and DFAD), PRE-DFKD is more robust for validation set-agnostic distillation with higher expectation and lower variance. %Although, DFAD can perform better than PRE-DFKD for CIFAR10, its performance is not consistent across other datasets.
When compared with CMI and MB-DFKD that store generated samples for replay, PRE-DFKD typically is at least similar and more often better. It might be counter-intuitive that although CMI stores and replays all generated samples, PRE-DFKD outperforms it. This is because the accuracy gain rate of the student models distilled via CMI is relatively low (see Figure \ref{fig:acc_plots}) causing lower expected value. We believe the gain rate is low because the novel samples constitute only a small proportion when compared to the replay samples. Lastly, MB-DFKD performing lower than PRE-DFKD suggests that limited amount of stored samples in MB-DFKD is not as effective as modeling the entire distribution of earlier observed samples in PRE-DFKD.
\par
Furthermore, to visualize the effect of pseudo replay on student accuracy throughout distillation, we plot the student accuracy curves (see Figure \ref{fig:acc_plots}) for the MNIST and CIFAR100 benchmarks. %\tm{need better explanation. Put MNIST and CIFAR100 in the graphs. Why choose these two? Why two different kinds of graphs?} 
For all methods, the observed accuracy series ($acc^{(i)}_{ts}$) across different runs on MNIST benchmark, had relatively higher variation than on other benchmarks. Therefore, for MNIST, we plot the accuracy curves with mean and variances across runs are indicated. For CIFAR100, we only display the mean accuracy curve ($\mu$) across four runs for each baseline.
\paragraph{Reduced Memory Footprint:}
We next compare PRE-DFKD with baselines using replay in terms of memory footprint. The additional memory required for the replay are noted in Table \ref{table:main}. Although CMI failed in Tiny ImageNet experiments, we report the memory it utilized after running for the same number of epochs (500) as other methods. The Table \ref{table:main} shows that PRE-DFKD significantly reduces the memory footprint for replay from hundreds of megabytes and even gigabytes to a constant 2MB, that is the memory occupied by the VAE parameters. Such reduction becomes more significant compared to CMI as the number of distillation steps increases since more batches of samples are generated and stored. The calculated memory footprint of CMI for different choices of distillation steps $ts$ and synthetic data dimensions are marked in Table \ref{table:CMI}. As the VAE architecture is independent of the selection of teacher-student pairs and distillation steps, the memory overhead does not scale based on them.  %\tm{Table 2 is not properly formatted. You should have a row called nt}
Although MB-DFKD maintains fixed-sized storage, the required storage for optimal distillation largely depends on the dataset. If the storage is small relative to the number of distillation steps, the effectiveness of replay diminishes. %\tm{don't you want to report distillation wall clock time}
\begin{table}[!h]
\begin{tabular}{l|l|l|l|l}
pixel dim. \textbackslash $ts$ & 2000   & 4000  & 8000   & 16000  \\ \hline
32                           & 2.5 GB & 5 GB  & 10 GB  & 20 GB  \\ \hline
64                           & 10 GB  & 20 GB & 40 GB  & 80 GB  \\ \hline
128                          & 40 GB  & 80 GB & 160 GB & 320 GB
\end{tabular}
\caption{CMI memory footprint for various pixel dimensions and distillation epochs}
\label{table:CMI}
\vspace{-15pt}
\end{table}
%the stochastic sample replacement procedure can cause ill-conditioned groups of replay samples to be contained  

\begin{figure}[!h]
    \centering
    %\includesvg[height=.18\textwidth]{figures/ablation.svg}
    \includegraphics[height=.18\textwidth]{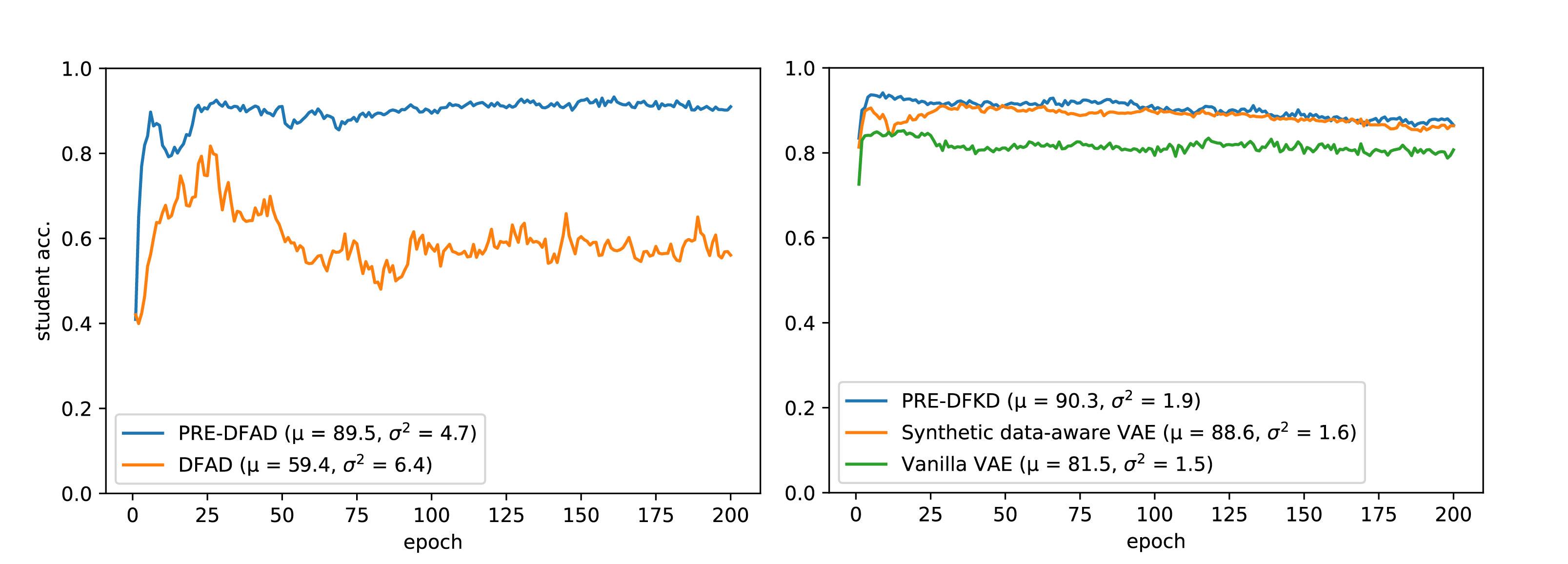}
    %\hspace{-15pt}
    %\includegraphics[height=.165\textwidth]{figures/PRE-DFKD_ablation.png}

    \caption{Ablation experiment results on MNIST.}
    \label{fig:ablation}
\end{figure}

%Table \ref{} suggests that, when compared with baselines that does not contain replay, our method is more reliable for validation set-agnostic distillation. This is inferred by the recorded higher expectation and lower variance of $acc_{nt}$. Although, DFAD can perform similar to replay-based methods in certain benchmarks, its performance is not consistent across datasets and runs (MNIST). When compared with CMI and MB-DFKD that store generated samples for replay, our method typically performs similar and even better. It might be unintuitive that although CMI stores and rehearses all generated samples, PRE-DFKD outperforms it. Our explanation for this is that the accuracy gain rate of student models distilled via CMI is relatively low (see Fig \ref{}) causing lower expected value. We believe the reason for the gain rate being low is the small proportion that novel samples have when compared to replay samples. 
\paragraph{Ablation Study:}
We now validate that the pseudo replay is effective regardless of the novel sample synthesis strategy. To do this, we couple our memory generator to the DFAD baseline method. In Figure \ref{fig:ablation}, DFAD distilled student accuracy is plotted with and without the inclusion of our memory generator. We name the VAE-enhanced baseline PRE-DFAD. For a fair comparison, we fix the random seeds for all the runs. The plots indicate that pseudo replay succeeded in preventing accuracy degradation caused by catastrophic forgetting even for DFAD. Additionally, we plot the PRE-DFKD results where the proposed class-balanced memory sample inference and synthetic data-aware reconstruction loss are omitted.
%In Fig. \ref{fig:ablation} DAFL and DFAD distilled student accuracy are plotted with and without the inclusion of our memory generator. We name the VAE-enhanced baselines as PRE-DAFL and PRE-DFAD respectively. For fair comparison, we fix the random seeds for all the runs. The plots indicate that pseudo replay succeeded in preventing accuracy degradation caused by catastrophic forgetting even for DAFL and DFAD. Additionally, we plot the PRE-DFAD results where the proposed class-balanced memory sample inference and synthetic data-aware reconstruction loss are omitted.
This way, we examine the individual contributions of these techniques. In Figure \ref{fig:ablation}, it is apparent that the exclusion of any of these impaired the student accuracy progressions. %\tm{this graph is very confusing. Why wouldn't you do the second part for PRE-DFKD? Moreover, how does PRE-DFKD compares with these enhanced versions? Is it still the best?}

\section{Conclusion}
In this work, we propose a data-free KD method that prevents the distilled model accuracy from degrading over time by generative pseudo replay. 
%is more reliable than existing approaches in the absence of validation data. 
While our method is more robust than existing approaches in the absence of validation data, it also significantly reduces the memory footprint and improves privacy preservation.
%While, improving reliability, our method manages to reduce the memory utilization as well. 
We believe these contributions provide a step towards allowing a wider audience to benefit from data-free KD, i.e., democratize it. In the future, our approach can be further improved by exploring ways to reduce the impact of dataset-specific hyper-parameter selection on distillation performance. 
\section{Acknowledgement}
This work was supported by the A*STAR Computational Resource Centre through the use of its high-performance computing facilities and by the National Research Foundation, Singapore under its Competitive Research Programme Award
NRF-CRP23-2019-0003.
\bibliography{refs}

\end{document}